# Trap-Based Pest Counting: Multiscale and Deformable Attention CenterNet Integrating Internal LR and HR Joint Feature Learning

Jae-Hyeon Lee and *Chang-Hwan Son

Department of Software Science & Engineering, Kunsan National University

558 Daehak-ro, Gunsan-si 54150, Republic of Korea

*Corresponding Author

Phone Number: 82-63-469-8915; Fax Number: 82-63-469-7432

E-MAIL: changhwan76.son@gmail.com; cson@kunsan.ac.kr

**Abstract**

Pest counting, which predicts the number of pests in the early stage, is very important because it enables rapid pest control, reduces damage to crops, and improves productivity. In recent years, light traps have been increasingly used to lure and photograph pests for pest counting. However, pest images have a wide range of variability in pest appearance owing to severe occlusion, wide pose variation, and even scale variation. This makes pest counting more challenging. To address these issues, this study proposes a new pest counting model referred to as *multiscale and deformable attention CenterNet* (Mada-CenterNet) for internal low-resolution (LR) and high-resolution (HR) joint feature learning. Compared with the conventional CenterNet, the proposed Mada-CenterNet adopts a multiscale heatmap generation approach in a two-step fashion to predict LR and HR heatmaps adaptively learned to scale variations, that is, changes in the number of pests. In addition, to overcome the pose and occlusion problems, a new between-hourglass skip connection based on deformable and multiscale attention is designed to ensure internal LR and HR joint feature learning and incorporate geometric deformation, thereby resulting in an improved pest counting accuracy. Through experiments, the proposed Mada-CenterNet is verified to generate the HR heatmap more accurately and improve pest counting accuracy owing to multiscale heatmap generation, joint internal feature learning, and deformable and multiscale attention. In addition, the proposed model is confirmed to be effective in overcoming severe occlusions and variations in pose and scale. The experimental results show that the proposed model outperforms state-of-the-art crowd counting and object detection models.

**1. Introduction**

Pest images captured by a light trap have a wide range of variability in pest appearance owing to severe occlusion and pose variation, and even scale variation. In particular, when a large number of pests are caught in a trap, they stick to each other and are obscured by other pests. The posture of pests changes in various ways. For example, their wings may be folded or unfolded. Pest wings can be clipped, distorting their shape. Moreover, pests appear similar because they have similar textures and colors. The number of pests can vary significantly. These issues make it difficult to distinguish them. Fig. 1 shows an example of pest images captured using a light trap. The pest counting problem, which aims to predict the number of pests from a pest image, is extremely challenging because of pose variation, changes in the number of pests, occlusion, and similar appearance in color and texture.

Two approaches can be considered for pest counting [1]. One is object detection that localizes bounding boxes from the pest image, and the other is crowd counting that predicts a density map to determine the number of objects in an image. Fig. 2 shows an example of the detected bounding boxes with an object detector and the density map predicted by a crowd counter. Thus far, these two approaches have been applied separately, depending on the number of objects, such as cars and pedestrians. In the case of a significantly large number of objects, crowd counting is used; in the opposite case, object detection is chosen. However, the number of pests varies greatly, as shown in Fig. 1. This raises the question of which approach is more suitable for pest counting. To the best of our knowledge, little research has been conducted on this topic thus far.

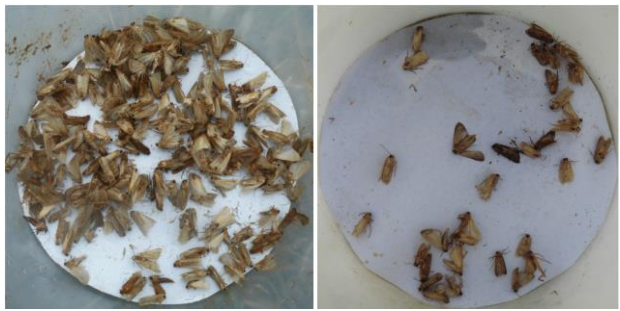

Fig. 1. Example of pest images captured by a light trap. Pose variation, occlusion, shape distortion, change in the number of pests, and similar appearance in color and texture make it difficult to distinguish pests. Therefore, pest counting problem is extremely challenging.

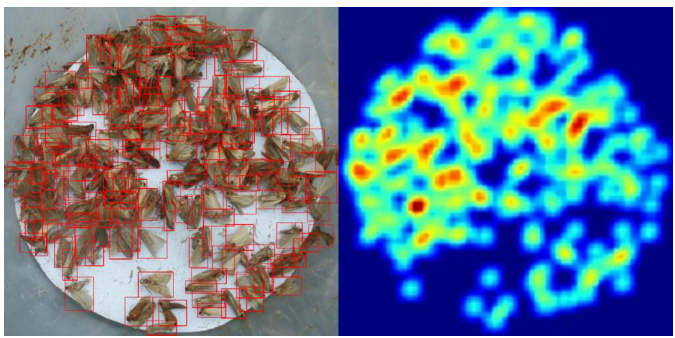

Fig. 2. Object detection (left image) vs. crowd counting (right image). Which approach is better for pest counting? Thus far, little research has been conducted on this.

**1.1. Proposed Mada-CenterNet vs. conventional CenterNet**

To address the aforementioned practical issues, this study introduces a new Mada-CenterNet, an advanced version of the conventional CenterNet [2], for pest counting. The first reason for choosing CenterNet as a base model for pest counting is that it is viewed as a hybrid approach combining bounding box localization and density map generation. Unlike existing object detectors, such as Faster R-CNN [3] and RetinaNet [4], that focus on predicting parameters related to bounding boxes via a regression function, CenterNet additionally exploits heatmaps that contain white colors to indicate the centroids of pests. Notably, heatmap generation is similar to the density map generation widely used for crowd counting. The minor difference is that the centroids of pests still have white colors to maintain the peak values for more accurate localization after Gaussian filtering. This heatmap generation can alleviate severe occlusions and wide pose variation problems more robustly than conventional object detectors. Therefore, hypothetically, the hybrid approach, CenterNet, is more suitable for pest counting than other object detection and crowd counting models. The second reason is that CenterNet is reported to outperform state-of-the-art object detection models in terms of speed and accuracy for object detection datasets (For example, COCO dataset [5]). However, the following aspects of CenterNet must be revised for pest counting.

- First, CenterNet predicts only a single-scale heat map. However, as shown in Fig. 1, the number of pests can vary significantly, depending on the timing of pest outbreaks. Large variations in the number of pests can cause scale problems. That is, when the number of pests is small, a small-scale heatmap is more effective and sufficient for pest counting, and vice versa, a large-scale heatmap is required to handle severe occlusion and wide pose variation problems. Therefore, CenterNet must adopt a multiscale heatmap generation approach. To this end, in the proposed Mada-CenterNet, low-resolution (LR) and high-resolution (HR) backbones are constructed for small-scale guided heatmap generation in a two-step fashion.
- Second, CenterNet uses stacked hourglasses as the backbone. However, information does not flow between the stacked hourglasses. The internal LR and HR features produced inside the hourglasses are hypothesized to be jointly learned, thereby boosting the discriminative power of the hourglasses. In the proposed Mada-CenterNet, a new between-hourglass skip connection is designed based on deformable and multiscale attention to transfer internal LR feature information to the HR hourglass. This approach helps to generate more accurate HR heatmaps and increase the pest counting accuracy. In other words, a new LR and HR joint feature learning is proposed for Mada-CenterNet.

- Third, because CenterNet was developed for object detection datasets with mild pose variation and occlusion, it excludes geometric transformation. However, as shown in Fig. 1, pest images can exhibit large pose variations and severe occlusions. To address these problems, the conventional CenterNet should consider geometric transformation to enhance the internal LR and HR features. In the proposed Mada-CenterNet, deformable convolution is newly adopted in the between-hourglass skip connection to be applied to the internal LR features that are jointly learned with internal HR features through multiscale attention, thereby focusing on more attentive areas and boosting the joint feature learning for more accurate pest counting.

## 2. Related Works

Pest counting techniques can be broadly classified into two categories [1]. One is crowd counting and the other is object detection. Crowd counting, which is also referred to as counting approaches by regression (or density estimation), predicts the density map where all pixels sum up to the number of pests, and object detection localizes the bounding boxes surrounding the pests. The number of detected bounding boxes is identical to that of pests in the image. This is the main difference between crowd counting and object detection methods. Fig. 3 shows examples of the predicted density map and detected bounding boxes. In the upper-right image, the generated density map is shown, where red indicates higher density areas and dark blue indicates lower density areas. In the density map, the sum of all pixels is equal to the number of pests. In the bottom-right image, the detected bounding boxes overlap on the input image.

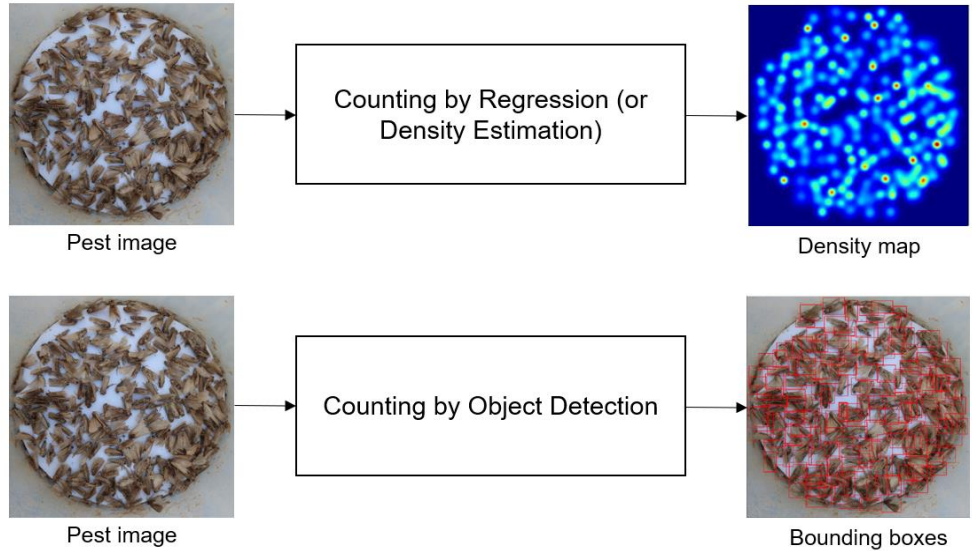

Fig. 3. Pest counting using density estimation and object detection approaches.

**2.1. Density estimation approach for pest counting**

In crowd images, people are too close and obscured by others. Therefore, traditional object detection based on bounding box localization is unsuitable for crowd counting. Hence, counting approaches using regression and density estimation have been developed. Classical crowd counting approaches [1] treat the counting problem as a regression problem to directly predict the crowd number from handcrafted image features, or indirectly estimate the crowd number from the generated density map. Recently, counting via density estimation using deep convolutional neural networks (DCNN) has become mainstream. Starting with CrowdNet [6], its multiscale and refinement variants such as MCNN [7], SANet [8], and ICCNet [9], which utilize different kernels and density maps in size, are introduced. More recently, kernel-based density map generation (KDMG) [10] that learns optimal kernels and updates the density map adaptively has shown outstanding performance.

**2.2. Object detection approach for pest counting**

The object detector determines the locations of the objects in the image. In general, two types of object detectors are used. One is a sliding window approach, and the other is a region proposal approach. The sliding window approach moves a sliding window along a raster scanning direction and determines whether an object is contained in the window, whereas the region proposal approach generates a thousand bounding box candidates while merging super-pixels based on similarity in a bottom-up grouping manner. Handcrafted features, such as histogram of gradients (HOG) [11] and scale invariant feature transform (SIFT) [12], are extracted, and a support vector machine (SVM) [13] is trained to classify the candidates. Popular region proposals include Selective Search [14] and Edge Boxes [15]. However, a recent trend is to use DCNN-based object detectors when large amounts of training data are available. DCNN-based object detectors can be divided into YOLO and RCNN families. The YOLO family is a single-shot detector that directly localizes bounding boxes from a full image using a regression model. Popular models include YOLO [16] and SSD [17]. The RCNN family is a two-stage detector that requires both region proposal and classifier learning. After RCNN, which is the earliest model, more advanced versions such as Faster RCNN [3], Mask RCNN [18], and Cascaded RCNN [19] have been developed.

## 2.3 Trap-based pest counting approach

Although pest and fruit counting methods such as MagoNet [20] and InsectNet [21] exist, in this study, we concentrate on *trap-based* pest counting approaches. This is because most fruit and leaf image datasets have a simple background or mild occlusion and a narrow variation in their shape, which is considerably different from our pest datasets, as already discussed with Fig. 1. With the advent of deep learning, moth detection [22] using a sliding window and trained classifier was introduced. Although this approach uses the DCNN, it still adopts a sliding window approach, and end-to-end learning is not realized for moth counting. Moreover, patch-based detection was conducted, implying that it is vulnerable to pest scale problems. Recently, PestNet [23] was introduced to first deal with large-scale pest detection, where channel and spatial attention are incorporated into the backbone to boost deep features, and a region proposal network is used with a position-sensitive score map to encode position information. However, the main architecture of PestNet is borrowed from Faster RCNN [3]. Our experiment used hybrid counting approaches and confirmed that the performance of Faster RCNN was inferior to that of CenterNet. Therefore, a more advanced pest counting model is required.

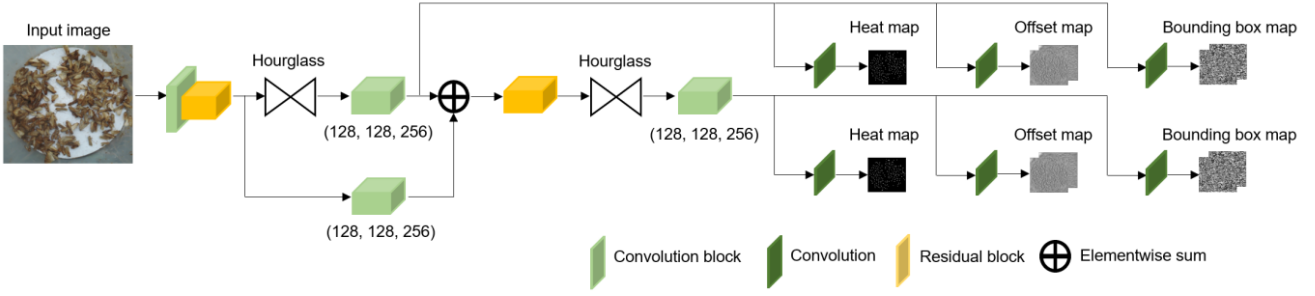

Fig. 4. Architecture of the conventional CenterNet for pest counting.

## 3. Background

Because the proposed Mada-CenterNet is regarded as an advanced version of the conventional CenterNet [2], an introduction to the background of CenterNet is necessary. CenterNet has proven powerful performance for object detection. Furthermore, it outperforms state-of-the-art object detection models such as Faster RCNN and RetinaNet in terms of speed and accuracy. Fig. 4 demonstrates the architecture of the CenterNet model. In Fig. 4, the CenterNet model uses two hourglasses as backbones for feature extraction and predicts three types of maps: heatmaps, offset maps, and bounding box maps. The two hourglasses are arranged in series and have the same scales in the feature domain. Unlike the conventional two-stage and single-shot object detectors, CenterNet requires to additionally predict two heatmaps with the same scale that are filled with white colors to indicate the centroids of the objects in the input image. Indeed, the centroids of the objects are identical to those of the bounding boxes that surround them. The centroid is referred to as the *keypoint* in [2].

$$\mathbf{I} \in \Re^{W \times H \times 3} \tag{1}$$

$$\mathbf{Y} \in [0,1]^{(W/R) \times (H/R) \times C} \tag{2}$$

where $\mathbf{I}$ denotes the input pest image, and $\mathbf{Y}$ denotes the heatmap. $W$ and $H$ denote the width and height of the input image, respectively. $R$ denotes the stride used to determine the resolution of the heatmap, and $C$ denotes the number of object classes. **Y** has a value of one for the keypoints and zero for other pixel locations.

$$\mathbf{Y} = \sum_{k\in\Omega} \delta(x - x_k) \otimes G \tag{3}$$

Gaussian filtering is applied at the key points to smooth the heatmap (**Y**), according to Eq. (3). Here, $\delta$ denotes the delta function, and $G$ denotes the Gaussian filter. $\otimes$ denotes the convolution operation, and $k$ denotes the keypoint. The generation of the heatmap is similar to that of the density map, which has been widely used for crowd counting [7]. However, a significant difference is observed between them. Compared with the density map, the pixel values at the key points in the heat map, which correspond to the white colors, remain unchanged after Gaussian filtering. Therefore, the summation of the heatmap is not equal to the number of pests in the pest image. The purpose of using the heatmap is to localize the centroids of the objects in the pest image. Therefore, the peak values should be maintained to effectively determine the keypoints with the brightest colors.

To train the CenterNet, a total loss function is defined as follows:

$$\mathcal{L} = \mathcal{L}_h + \lambda_b \mathcal{L}_b + \lambda_o \mathcal{L}_o \tag{4}$$

$$\mathcal{L}_h = -\frac{1}{N}\sum_{k}^{N} \begin{cases} (1-\hat{\mathbf{Y}}(k))^\alpha \log(\hat{\mathbf{Y}}(k)) & if\ \mathbf{Y}(k)=1 \\ (1-\mathbf{Y}(k))^\beta (\hat{\mathbf{Y}}(k))^\alpha \log(1-\hat{\mathbf{Y}}(k)) & otherwise \end{cases} \tag{5}$$

$$\mathcal{L}_b = \frac{1}{N}\sum_{k=1}^{N} |\hat{\mathbf{S}}_k - \mathbf{S}_k| \tag{6}$$

$$\mathcal{L}_o = \frac{1}{N}\sum_{k}^{N} \left|\hat{\mathbf{O}}_{\tilde{k}} - \left(\frac{k}{R} - \tilde{k}\right)\right|,\ where\ \ \tilde{k} = \left[\frac{k}{R}\right] \tag{7}$$

$\mathcal{L}_h$, $\mathcal{L}_b$, and $\mathcal{L}_o$ calculate the prediction errors for the ground-truth heatmap, bounding box map, and offset map. First, to model a loss function for the heatmap, the focal loss, which is a variant of cross entropy, is used to address the class imbalance problem during training, as shown in Eq. (5). Here, $\hat{\mathbf{Y}}$ denotes the predicted heatmap, and a prediction $\mathbf{Y}(k) = 1$ corresponds to the $k$-th key point in the ground-truth heatmap. $N$ denotes the total number of keypoints in the input image. The focal loss downweights the loss for well-classified examples and focuses more on difficult, misclassified examples. In Eq. (5), $\alpha$ and $\beta$ are set to 2 and 4, respectively. Second, in Eq. (6), $\mathbf{s}_k$ contains the width and height of the ground-truth bounding box at the $k$-th keypoint, and $\hat{\mathbf{S}}_{p_k}$ is the predicted bounding box map that has the same size as $\hat{\mathbf{Y}}$, but the number of channels is two. Thus, $\mathcal{L}_b$ is the sum of the errors between the predicted and ground-truth bounding boxes. Third, $\mathcal{L}_o$ is required to reflect the discretization errors caused by downsampling at a ratio of $R$. In Eq. (7), parenthesis implies rounding off to obtain an integer pixel location, and $\hat{\mathbf{O}}_{\tilde{p}}$ denotes the offset map that has the same size as $\hat{\mathbf{S}}_{p_k}$ and contains offsets for 2D pixel coordinates. In Eq. (4), $\lambda_b$ and $\lambda_o$ denote weights that are set to 1 and 0.1, respectively. To reduce the total loss in Eq. (4) iteratively, gradient-based optimizers [24] can be used.

In the test phase, max pooling is first applied to the heatmap, which is predicted by the latter hourglass, to remove noise and determine the keypoints with the brightest colors. Subsequently, at the keypoints, bounding boxes are detected using the offset and bounding box maps. Therefore, in the case of CenterNet, locating the keypoints is crucial, resulting in an increase in pest counting accuracy.

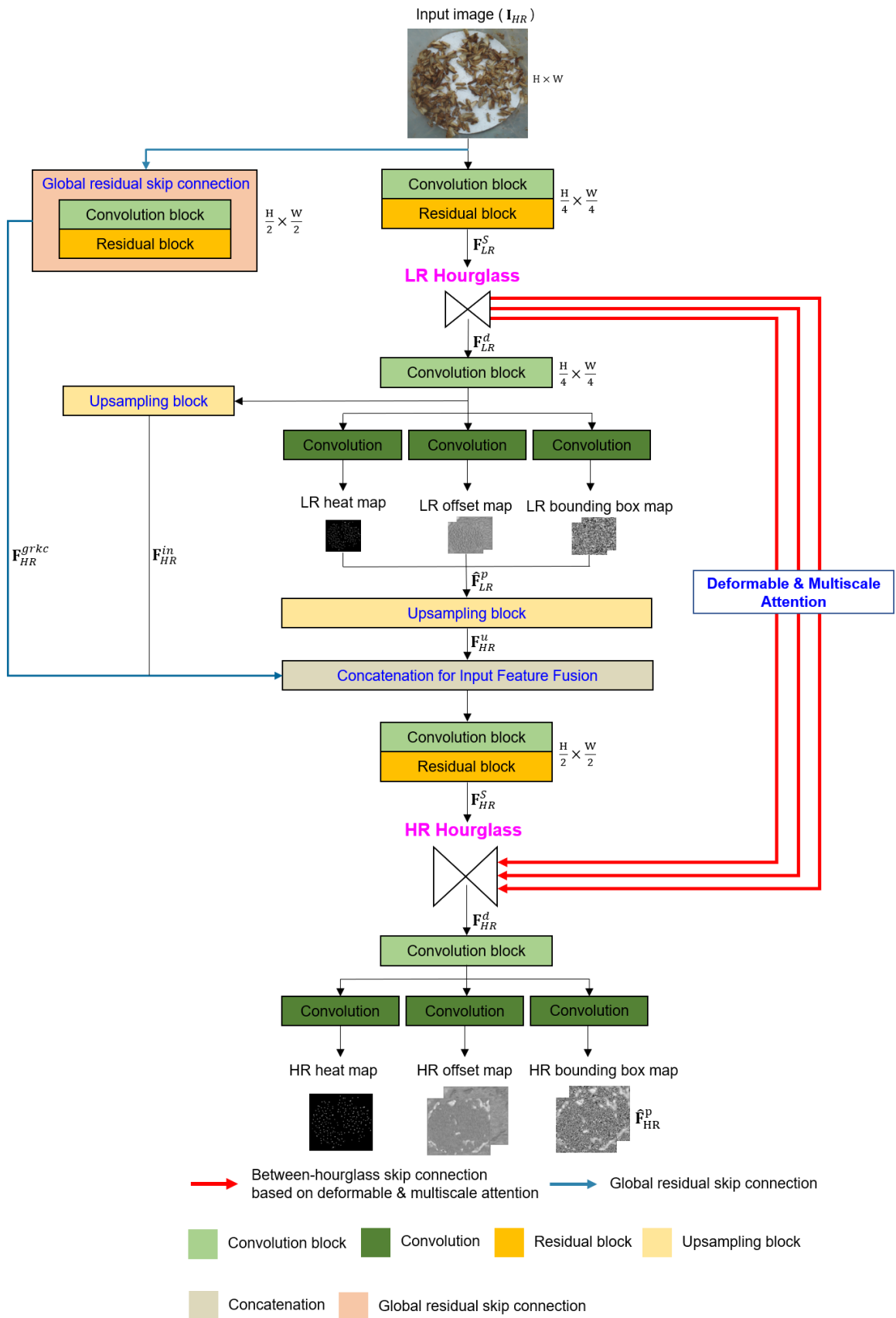

Fig. 5. Proposed Mada-CenterNet for pest counting.

## 4. Proposed Mada-CenterNet for Internal Multiscale Joint Feature Learning

The conventional CenterNet has certain drawbacks. As shown in Fig. 4, the two hourglasses have the same scale in the feature domain. In addition, feature information does not flow between them. Moreover, geometric deformation is not considered. In the case of pest counting, severe occlusion, wide pose variation, and changes in the number of pests can appear in pest images, as shown in Fig. 1. To cope with these problems, conventional CenterNet must be advanced. Thus, a multiscale and deformable model based on internal multiscale joint feature learning is required for a more accurate pest counting.

The architecture of the proposed Mada-CenterNet to incorporate deformable and multiscale attention based on internal LR and HR multiscale joint feature learning is illustrated in Fig. 5; noticeable differences exist compared with Fig. 4. First, a between-hourglass skip connection, which is drawn with thick red lines in Fig. 5, is newly constructed. The internal LR features produced inside the LR hourglass flow into the HR hourglass via the between-hourglass skip connection to realize deformable and multiscale attention. This design enables the transfer of internal LR feature information into the HR hourglass and focuses on more important areas in the HR feature domain, thereby alleviating pose deformation and occlusion problems. Second, the internal multiscale features of the LR and HR hourglasses are extracted and fused in the proposed Mada-CenterNet. As shown in Fig. 5, LR and HR hourglasses are used as the LR and HR feature extractors, respectively. In pest images, the number of pests varied widely. In the case of a small number of pests, extracting small-scale features and predicting small-scale heat maps are more efficient and sufficient. In the opposite case, larger features are required owing to occlusion. Through the proposed between-hourglass skip connection, the LR hourglass provides small-scale internal features to the HR hourglass for multiscale-based attention, thereby boosting the discriminative power of the HR hourglass. In other words, the LR hourglass plays the role of a teacher network to transfer internal LR feature knowledge to the HR hourglass. Therefore, the proposed Mada-CenterNet can adapt to the number of pests in the input image, thereby alleviating the scaling problem and increasing the discriminative power of the HR hourglass. Third, geometric deformation is incorporated into the between-hourglass skip connection, where internal LR features are sampled to determine more discriminative LR features and jointly learn HR features, agnostic to pest occlusion and pose variation. This approach can enhance internal multiscale joint feature learning more effectively, thereby improving the pest counting accuracy.

The proposed Mada-CenterNet largely consists of an LR hourglass, upsampling feature transformation, global residual skip connection and input feature fusion, between-hourglass skip connection based on deformable and multiscale attention for internal multiscale joint feature learning, and an HR hourglass.

### 4.1. Prediction of LR maps

The input pest image is first embedded into the LR feature domain through the convolution and residual blocks and then fed into the LR hourglass for deep feature extraction.

$$\mathbf{F}_{LR}^{s} = f_{RB} \circ f_{CB}(\mathbf{I}_{HR}) \tag{8}$$

$$\mathbf{F}_{LR}^{d} = \mathcal{H}_{LR}(\mathbf{F}_{LR}^{s}) \tag{9}$$

where $\mathbf{I}_{HR}$ denotes the pest color image, and $f_{RB}$ and $f_{CB}$ denote the residual and convolution blocks, respectively. Symbol ° is a composite function, and $\mathcal{H}_{LR}$ denotes the LR hourglass. $\mathbf{F}_{LR}^{s}$ and $\mathbf{F}_{LR}^{d}$ correspond to shallow and deep features, respectively, before and after passing through the LR hourglass.

To map deep features $\mathbf{F}_{LR}^{d}$ to LR predictions, three types of maps, one convolution block, and one convolution operation are additionally applied.

$$\hat{\mathbf{F}}_{LR}^{p} = \{\hat{\mathbf{F}}_{LR}^{H}, \hat{\mathbf{F}}_{LR}^{O}, \hat{\mathbf{F}}_{LR}^{B}\} = f_{C} \circ f_{CB}(\mathbf{F}_{LR}^{d}) \tag{10}$$

$\hat{\mathbf{F}}_{LR}^{H}$, $\hat{\mathbf{F}}_{LR}^{O}$, and $\hat{\mathbf{F}}_{LR}^{B}$ correspond to the predicted LR heatmap, offset map, and bounding box map, respectively. To train the three types of LR maps, a new LR loss function is defined as follows:

$$\mathcal{L}(\widehat{\mathbf{F}}_{LR}^p) = \mathcal{L}(\widehat{\mathbf{F}}_{LR}^H) + \lambda_{LR}^o \mathcal{L}(\widehat{\mathbf{F}}_{LR}^O) + \lambda_{LR}^b \mathcal{L}(\widehat{\mathbf{F}}_{LR}^B) \qquad (11)$$

$$\mathcal{L}(\widehat{\mathbf{F}}_{LR}^H) = -\frac{1}{N}\Sigma_k \begin{cases} \left(1 - \widehat{\mathbf{F}}_{LR}^H(k)\right)^\alpha log\left(\widehat{\mathbf{F}}_{LR}^H(k)\right) & if\ \mathbf{F}_{LR}^H(k) = 1 \\ \left(1 - \widehat{\mathbf{F}}_{LR}^H(k)\right)^\beta \left(\widehat{\mathbf{F}}_{LR}^H(k)\right)^\alpha log(1 - \widehat{\mathbf{F}}_{LR}^H(k)) & otherwise \end{cases} \qquad (12)$$

$$\mathcal{L}(\widehat{\mathbf{F}}_{LR}^B) = \frac{1}{N}\Sigma_k \left|\widehat{\mathbf{F}}_{LR}^B(k) - \mathbf{F}_{LR}^B(k)\right| \qquad (13)$$

$$\mathcal{L}(\widehat{\mathbf{F}}_{LR}^O) = \frac{1}{N}\Sigma_k \left|\widehat{\mathbf{F}}_{LR}^O(k) - \left(\frac{k}{R} - \tilde{k}\right)\right|,\ where\ \ \tilde{k} = \left[\frac{k}{R}\right] \qquad (14)$$

Here, $\widehat{\mathbf{F}}_{LR}^B$ stores the width and height of the bounding box map at the key point, and $\widehat{\mathbf{F}}_{LR}^O$ has offset information. $\widehat{\mathbf{F}}_{LR}^B$ and $\widehat{\mathbf{F}}_{LR}^O$ have the same number of channels, that is, two. $\widehat{\mathbf{F}}_{LR}^H$ is a grayscale map because the pest captured by the trap includes only one species.

To generate the ground-truth LR heatmap, the HR bounding boxes are first scaled down according to stride $R$, and then LR keypoints are redefined. The newly rendered LR heatmap has white colors at the redefined keypoints. Subsequently, Gaussian filtering is applied to blur the LR heatmap, according to Eq. (3). Here, notably, the white colors are not unchanged after Gaussian filtering to maintain the peaks, enabling easy determination of the keypoints. This is the main difference between the heatmap and the density map. An offset map is created using discretized centroids of LR bounding boxes.

### 4.2. Upsampling feature transformation

The deep features output by the last convolution block behind the LR hourglass and the three predicted types of maps are exploited to predict the HR maps more accurately. However, a scale mismatch exists between the LR and HR maps. Therefore, feature scaling must be performed.

$$\mathbf{F}_{HR}^u = f_{UB}\left(\widehat{\mathbf{F}}_{LR}^p = \{\widehat{\mathbf{F}}_{LR}^H, \widehat{\mathbf{F}}_{LR}^O, \widehat{\mathbf{F}}_{LR}^B\}\right) \qquad (15)$$

$$\mathbf{F}_{HR}^{in} = f_{UB} \circ f_{CB}(\mathbf{F}_{LR}^{d}) \tag{16}$$

Here, $f_{UB}$ denotes the upsampling block consisting of upsampling and convolution layers used to enlarge the feature maps. In this study, bicubic interpolation is used to implement the upsampling layer.

### 4.3. Global residual skip connection and input feature fusion

A better approach would be for the HR hourglass to utilize the information about the input pest images. To this end, a global residual skip connection (GRKC) is considered. In this study, residual and convolution blocks are used to design GRKC, as shown in Fig. 5. Through GRKC, visual information of the input pest image can be transferred to the HR hourglass.

$$\mathbf{F}_{HR}^{s} = f_{RB} \circ f_{CB} \circ f_{CONCAT}(\mathbf{F}_{HR}^{in}, \mathbf{F}_{HR}^{up}, \mathbf{F}_{HR}^{grkc}) \tag{17}$$

$\mathbf{F}_{HR}^{grkc}$ represents the output feature map of GRKC, and $f_{CONCAT}$ represents the concatenation for feature fusion. In Eq. (17), the upsampled LR feature maps, including the LR heatmap, offset map, and bounding box map, are fused with the input pest image in the feature domain, making the input features richer and more discriminative.

### 4.4. Between-hourglass skip connection based on deformable and multiscale attention

Fused input features $\mathbf{F}_{HR}^{s}$ contain the three predicted types of LR maps, and the goal of the HR hourglass is to predict the corresponding HR maps from $\mathbf{F}_{HR}^{s}$. Therefore, fused input features $\mathbf{F}_{HR}^{s}$ help to improve the performance of HR map prediction. In addition, the architecture of the LR hourglass is the same as that of the HR hourglass. Only the sizes of the internal feature maps are different. Unlike the conventional CenterNet, in this study, the LR hourglass is connected to the HR hourglass via a *between-hourglass skip connec*tion, as shown by the red lines in Fig. 5. In other words, the internal LR features produced inside the LR hourglass are fed into the HR hourglass to be jointly learned with the internal HR features. To fuse the internal LR and HR features at different scales, deformable and multiscale attention is designed. Fig. 6 illustrates the detailed architecture of the proposed between-hourglass skip connection, built based on deformable convolution and multiscale attention for internal multiscale joint feature learning.

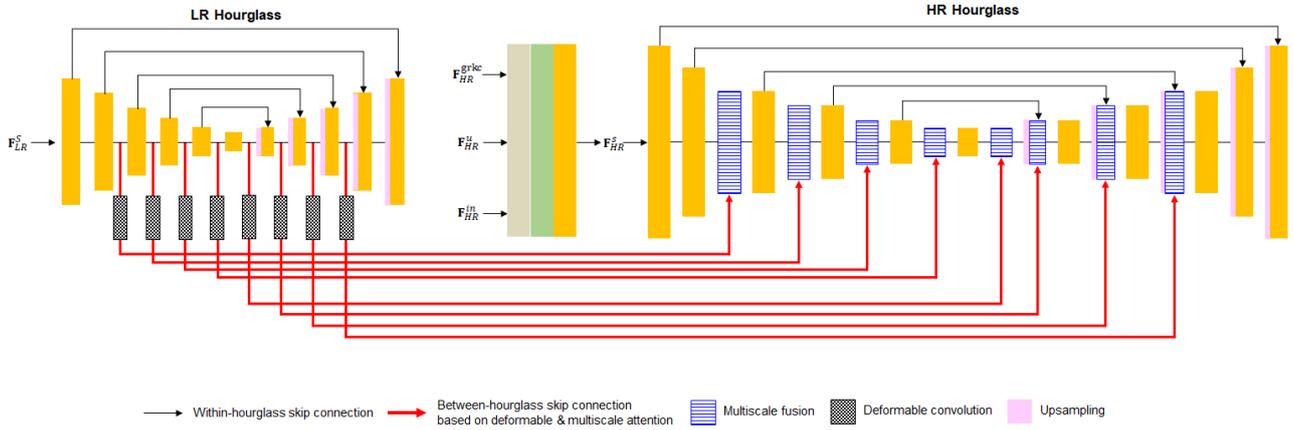

Fig. 6. Proposed between-hourglass skip connection based on deformable and multiscale attention for internal multiscale joint feature learning.

### 4.4.1. Internal LR feature deformation

The standard convolution extracts local features with many filters in the DCNN and has shown power performance for feature learning, particularly for computer vision problems. However, the standard convolution has an inherent limitation in modeling geometric deformation because it can only extract local features at regular grids from the center of the filter to be slid. To this end, deformable convolution was devised. Unlike the standard convolution, deformable convolution adds 2D offsets to regular grids in the standard convolution, thereby enhancing the DCNN capability of modelling geometric transformation.

The pest images captured in a light trap, which are targeted in this study, have severe occlusion and wide pose variation. Deformable convolution is considered to model geometric deformation. In the proposed Mada-CenterNet, deformable convolution is inserted into the between-hourglass skip connection to apply geometric deformation to the internal LR features, as shown in Fig. 6.
16

$$\mathbf{F}_{LR}^{\mathcal{H}(\ell)} = f_{DC}\left(\mathcal{H}_{LR}^{\ell}(\mathbf{F}_{LR}^{s})\right) \tag{18}$$

Here, $\mathcal{H}_{LR}^{\ell}$ indicates the output feature maps at the $\ell$-th residual block (RB) in the LR hourglass, and $f_{DC}$ denotes the deformable convolution. In the LR hourglass, $f_{DC}$ is not applied to the first and last two RBs because of the computational complexity during multiscale attention. Eq. (18) indicates that the LR hourglass produces the deformed version $\mathbf{F}_{LR}^{\mathcal{H}(\ell)}$ of internal LR feature map $\mathcal{H}_{LR}^{\ell}$. Deformed LR features $\mathbf{F}_{LR}^{\mathcal{H}(\ell)}$ are transferred to the HR hourglass for internal multiscale attention fusion.

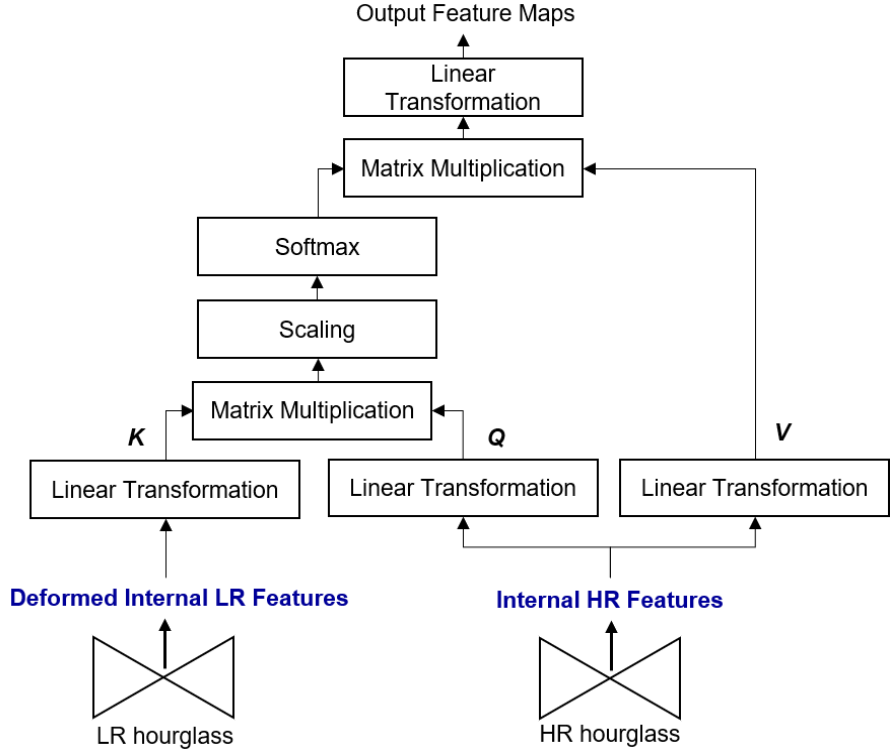

Fig. 7. Deformed internal LR feature guided multiscale attention

### 4.4.2. Internal LR feature guided multiscale attention

The details of multiscale attention for internal LR and HR feature fusion are shown in Fig. 7. Unlike the original attention model in Transformer [25], three types of inputs, key (K), query (Q), and value (V), are visual feature maps, and K has a different scale than Q and V. In other words, K is small-scale and takes linearly transformed internal LR features $\mathbf{F}_{LR}^{\mathcal{H}(\ell)}$ as the input. In addition, K represents deformed LR features. In contrast, Q and V are large-scale. The internal HR feature maps are assigned to Q and V after applying linear transformation. In this study, scaled dot-product attention [25] is chosen to implement multiscale attention. In Fig. 6, the HR hourglass follows the encoder–decoder framework, and thus, multiscale attention is implemented slightly differently depending on the encoder and decoder.

For the encoder of the HR hourglass, multiscale attention is designed as follows:

$$\boldsymbol{Q} = \boldsymbol{V} = \mathbf{F}_{HR}^{\mathcal{H}(\ell)} = \mathcal{M}\left(\mathcal{H}_{HR}^{\ell}(\mathbf{F}_{HR}^{S})\right) \tag{19}$$

$$\boldsymbol{K} = \mathcal{M}\left(\mathbf{F}_{LR}^{\mathcal{H}(\ell)}\right) \tag{20}$$

$$\text{Attention}^{encoder}(\mathbf{Q},\mathbf{K},\mathbf{V}) = \text{softmax}\left(\frac{\boldsymbol{Q}\boldsymbol{K}^T}{\sqrt{d_k}}\right)\boldsymbol{V} \tag{21}$$

where $\mathcal{H}_{HR}^{\ell}$ indicates the output feature maps at the $\ell$-th RB in the HR hourglass. Similar to the LR hourglass, the first two and last three RBs are excluded from the multiscale fusion. In Eq. (19), input feature map $\mathbf{F}_{HR}^{S}$ passes through the HR hourglass, followed by linear transformation $\mathcal{M}$ to produce internal HR features $\mathbf{F}_{HR}^{\mathcal{H}(\ell)}$ that are then assigned to Q and V. Similarly, deformed internal LR features $\mathbf{F}_{LR}^{\mathcal{H}(\ell)}$ are assigned to K after linear transformation $\mathcal{M}$, as shown in Eq. (20). Scaled dot-product attention is used to implement the multiscale attention of the encoder. In Eq. (21), $d_k$ denotes the dimension of K, and softmax represents the softmax function used to calculate weights between 0 and 1. Here, deformed internal LR features $\mathbf{F}_{LR}^{\mathcal{H}(\ell)}$ are used to calculate similarity matrix $\boldsymbol{Q}\boldsymbol{K}^T$ and determine the LR features that are more important for pest counting. In other words, it serves as a guide for learning weights for internal HR features $\mathbf{F}_{HR}^{\mathcal{H}(\ell)}$.

For the decoder of the HR hourglass, multiscale attention additionally requires the internal HR features transferred by the encoder. The multiscale attention for the decoder is modified as follows:

$$\text{Attention}^{decoder}(\mathbf{Q}, \mathbf{K}, \mathbf{V}) = \text{softmax}\left(\frac{\mathbf{Q}\mathbf{K}^T}{\sqrt{d_k}}\right)\mathbf{V} + \mathbf{S} \qquad (22)$$

Here, $\mathbf{S}$ represents the internal HR features transferred by the encoder via *within-hourglass skip connection*, as shown in Fig. 6, and then added to the scaled dot-product attention result.

In the proposed multiscale attention, internal LR features are jointly learned with internal HR features via *between-hourglass skip connection*. The internal LR features serve as a guide for enhancing the internal HR features. Compared with other Vision Transformers (VTs) [26,27], in this study, two types of internal LR and HR features, which are outputs of two backbones, are learned jointly based on scaled dot-product attention. Notably, the internal LR features are deformed to be more robust to pest occlusion and wide pose variation, enabling the internal HR features to be more discriminative. Other VTs use only one backbone; thus, the two types of internal multiscale features are not considered. This is a key difference between the proposed multiscale attention and other VTs.

### 4.5. Prediction of HR maps

Internal HR features can be made more discriminative through the proposed multiscale attention, where the deformed internal LR features are jointly learned with the internal HR features to focus on the more important areas in the feature domain. The output features of the HR hourglass pass through the convolution block and are then transformed into final prediction HR maps.

$$\hat{\mathbf{F}}_{HR}^{p} = \{\hat{\mathbf{F}}_{HR}^{H}, \hat{\mathbf{F}}_{HR}^{O}, \hat{\mathbf{F}}_{HR}^{B}\} = f_C \circ f_{CB}(\mathbf{F}_{HR}^{d}) \qquad (23)$$

Here, $\mathbf{F}_{HR}^{d}$ represents the output feature map of the HR hourglass, and $\hat{\mathbf{F}}_{HR}^{H}, \hat{\mathbf{F}}_{HR}^{O}$, and $\hat{\mathbf{F}}_{HR}^{B}$ correspond to the predicted HR heatmap, offset map, and bounding box map, respectively. To train the HR maps, a HR loss function is defined as follows:

$$\mathcal{L}(\hat{\mathbf{F}}_{HR}^p) = \mathcal{L}(\hat{\mathbf{F}}_{HR}^H) + \lambda_{HR}^o \mathcal{L}(\hat{\mathbf{F}}_{HR}^O) + \lambda_{HR}^b \mathcal{L}(\hat{\mathbf{F}}_{HR}^B) \qquad (24)$$

The HR loss function is the same as the LR loss function in Eq. (4), except that predicted LR maps $\hat{\mathbf{F}}_{LR}^p$ of the loss function is replaced with predicted HR maps $\hat{\mathbf{F}}_{HR}^p$. Once again, the white colors in the ground-truth HR heatmap remain unchanged after Gaussian filtering to maintain the peak values, enabling the easy determination of keypoints. This is the key difference between the heatmap and the density map.

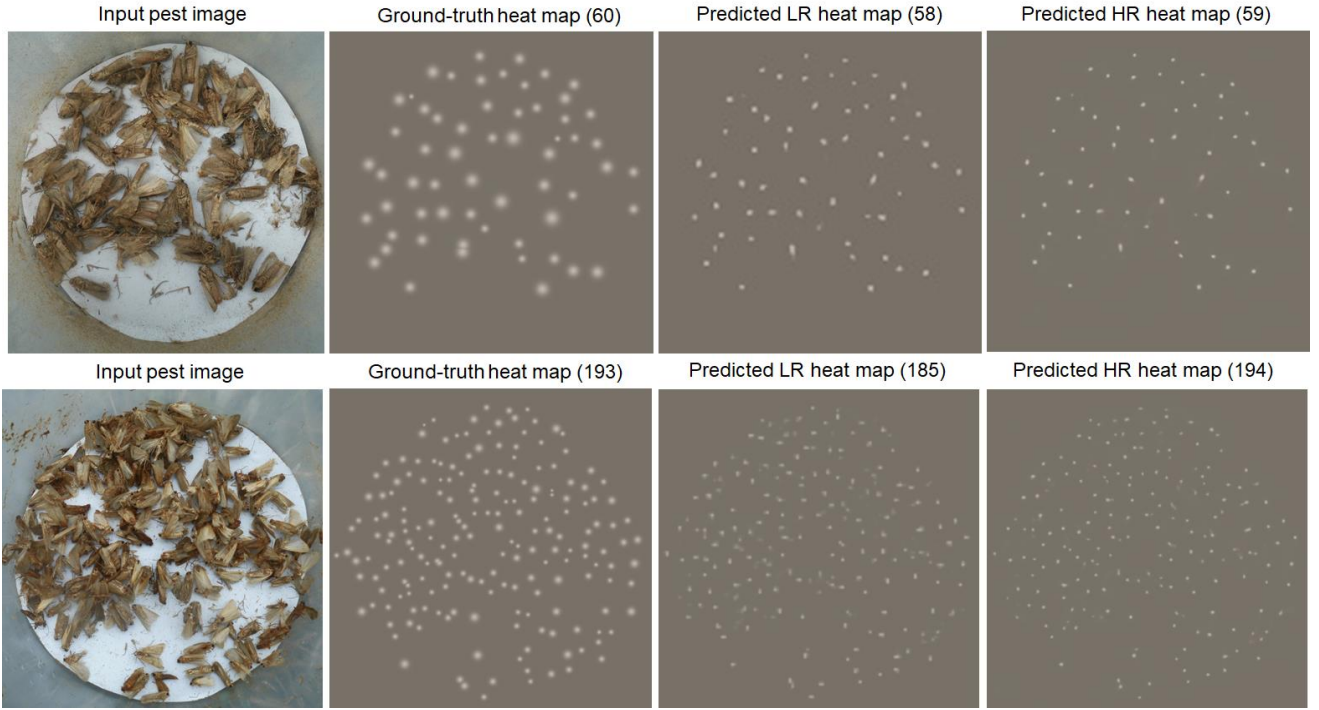

Fig. 8. Experiments for scale variation; input pest images, ground-truth heatmaps, predicted LR heatmaps, and predicted HR heatmaps (left to right).

## 5. Experimental Results

In this study, two types of pests that mainly harm soybean were captured by a light trap and used as training and test datasets for pest counting. One is Spodoptera exigua, and the other is Spodoptera litura. The total number of pests was 4462. The data augmentation technique, which includes flip, contrast, brightness, and saturation, was applied to generate synthetic pest images. The ratio of the training to test datasets was 7:3. Adam [24] was used as the optimizer, and the batch size was 4. The number of epochs was 100, and the learning rate was set to 0.001. PyTorch was used as the deep learning framework. Input pest images had a fixed size of 512 × 512, and stride $R$ was set to 4 for downsampling.

**5.1. Verification for multiscale approach**

To verify the effectiveness of the multiscale heatmap generation approach, LR and HR heatmaps were compared. Because these heatmaps include white colors at the key points, visually comparing the LR and HR maps is easy. However, other offset and bounding box maps were not compared because of the absence of texture patterns in those maps. Fig. 8 shows the ground truth and predicted LR and HR heatmaps. In Fig. 8, the numbers in parentheses above the heatmaps correspond to the number of pests. In Fig. 8, the LR heatmaps are resized with bicubic interpolation for visual comparison such that the image size is the same as that of the HR heatmaps. Owing to the upsampling effect, the LR heatmaps were blurry at the keypoints. As shown in the upper part of Fig. 8, where the number of pest images is relatively small, the predicted pest count from the LR heatmap was as accurate as that from the HR heatmap. This was possible because the occlusion was not severe. This indicated that small-scale LR heatmap predictions were sufficient for pest counting. However, in the opposite case, the accuracy of the LR heatmap was lower than that of the HR heatmap. This was because downsampling caused pixel information to be lost, exacerbating the occlusion problem. To address this scale issue, a multiscale approach was adopted in this study for LR and HR map prediction. In a two-step process, small-scale LR maps were first predicted and used as side information to generate the corresponding large-scale HR map more accurately.

To compare the accuracies of the predicted HR and LR heatmaps for the input test images, the absolute error (AE) was evaluated. The AE scores of the HR and LR heatmaps were calculated as follows:

$$AE_{i\in\{HR,LR\}} = \left|p(\hat{\mathbf{F}}_i^H) - p(\mathbf{F}_i^H)\right| \tag{25}$$

Here, $i$ indicates one of the HR and LR heatmaps, and $\hat{\mathbf{F}}_i^H$ and $\mathbf{F}_i^H$ represent the predicted and ground-truth heat maps, respectively. $p$ calculates the number of pests by localizing the keypoints and detecting the bounding boxes from the given heatmap. Therefore, the $AE$ is the absolute difference between the predicted and the measured number of pests. For reference, keypoint localization was conducted based on max pooling, and the bounding boxes were detected using the offset and bounding box maps. Further details on keypoint localization are described in [2].

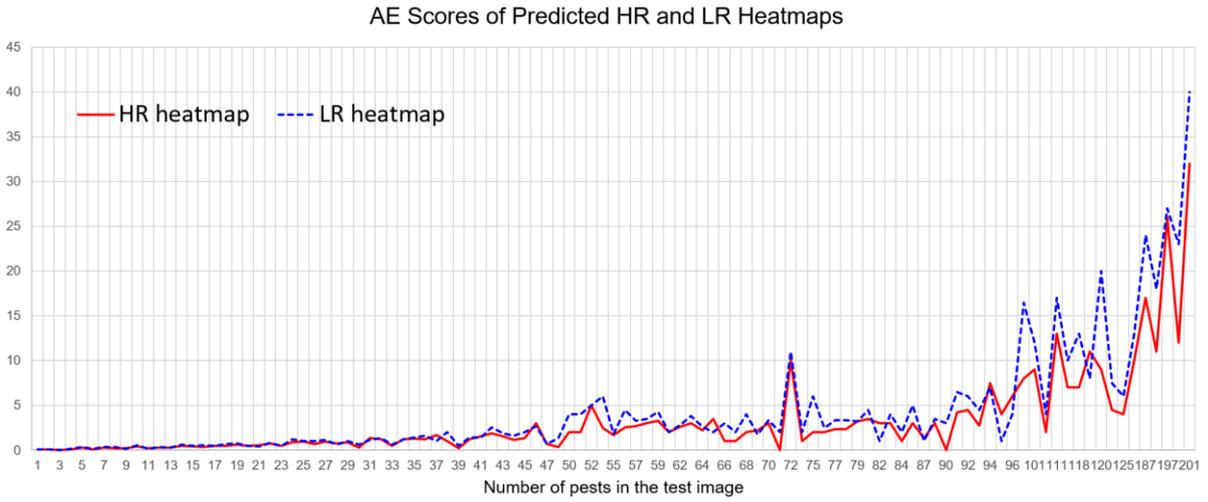

Fig. 9. AE scores of the predicted HR and LR heatmaps.

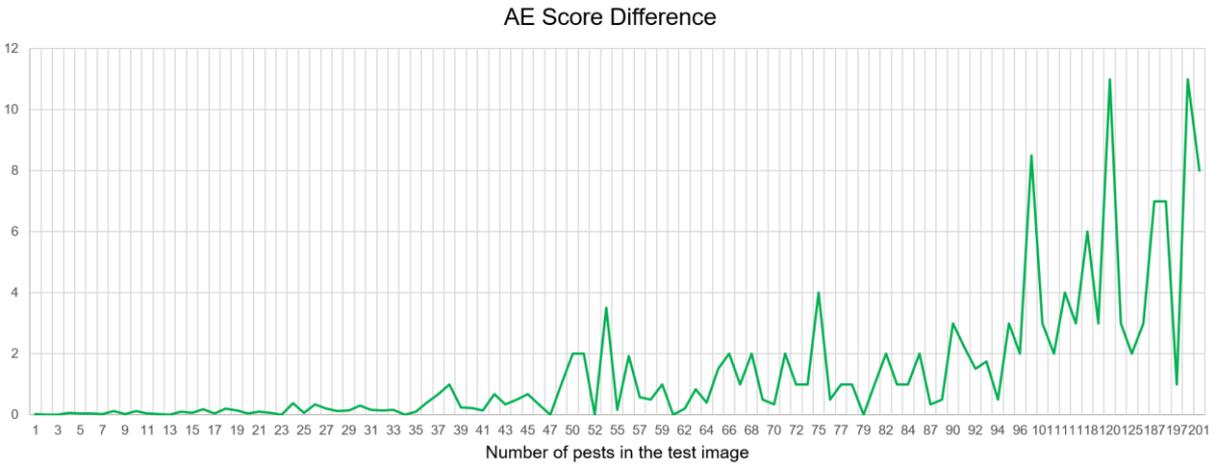

Fig. 10. AE score difference for HR and LR heatmaps.

Table 1. Quantitative evaluation for SOTA pest counting models

| Method | SOTA models | MAE (↓) | AP (↑) |
|---|---|---|---|
| Counting by density map estimation | SANet [8] | 2.114 | - |
| | MCNN [7] | 4.104 | - |
| | ICCNet [9] | 9.710 | - |
| | KDMG [10] | 1.273 | - |
| Counting by detection | RetinaNet [4] | 4.634 | 0.748 |
| | Faster RCNN [3] | 3.312 | 0.802 |
| | RepPoints [28] | 1.471 | 0.918 |
| | CenterNet [2] | 0.766 | 0.953 |
| | **Proposed Multiscale CenterNet** | **0.752** | **0.961** |
| | **Proposed Multiscale Attention CenterNet** | **0.711** | **0.963** |
| | **Proposed Mada-CenterNet** | **0.696** | **0.968** |

Fig. 9 shows the AE scores of the predicted HR and LR heatmaps according to the number of pests. When the number of pests was less than 90, the AE scores of the LR heatmaps were similar to those of the HR heatmaps. To be more specific, the AE score difference was calculated, as shown in Fig. 10. As shown in this figure, the AE score difference increased rapidly after 90. This indicated that the LR heat map was not suitable when the number of pests was large. That is, the performance of the LR heatmap deteriorated because of the occlusion problem. Therefore, when the number of pests increases, an HR heat map is required. This experiment confirmed that the proposed multiscale heatmap generation approach was reasonable. As shown in Fig. 9, the sum of the AE scores of the LR heatmaps was 451.2 and that of the HR heatmaps was 335.1. In other words, the accuracy of the HR heatmap was higher than that of the LR heatmap. This indicated that the use of the LR heatmap could be an effective way to improve the accuracy of the HR heatmap within a stacked hourglass framework. For reference, to check the effectiveness of only the proposed multiscale heatmap generation, the between-hourglass skip connection was removed (from the architecture shown in Fig. 6). The proposed model without the between-hourglass skip connection was tested for AE calculation. In addition, the AE scores were averaged over the test images with the same number of pests.

**5.2. Performance evaluation for pest counting models**

Two types of pest counting models were compared for quantitative evaluation. One is counting by detection and the other is counting by density map estimation. State-of-the-art (SOTA) models such as MCNN [7], SANet [8] and ICCNet [9], KDMG [10], RetinaNet [4], RedPoints [28], CenterNet [2], and Faster RCNN [3] were tested. Table 1 presents the mean absolute error (MAE) [7] and averaged precision (AP) [3] for the SOTA pest counting models. Notably, KDMG [10] outperformed the counting by detection models, except for CenterNet [2] and the proposed models, which are classified by counting by detection models. This result indicated that crowd counting models could be directly used for pest counting applications and could obtain excellent performance not only for pest images but also for crowd images. Pest images have severe pose variation and occlusion problems, preventing counting by detection models from determining bounding boxes more accurately. However, KDMG can learn kernel shapes and iteratively update the density map, thereby significantly improving the pest counting accuracy. Furthermore, KDMG has the best performance among conventional crowd counting models [10].

Although KDMG was the best among the tested SOTA models, its performance was inferior to that of CenterNet and the proposed models. Unlike conventional object detection models, CenterNet predicts a heatmap and then determines the keypoints for bounding box detection. This heatmap appears similar to the density map used for crowd counting models. This result reveals that the CenterNet-like architecture based on heatmap prediction can cope with pose variations and occlusion problems for accurate pest counting.

The proposed model outperformed all the tested SOTA models. In particular, the proposed model showed a better performance than CenterNet and KDMG. The proposed model is an advanced version of the CenterNet. Specifically, the proposed method revises the conventional CenterNet in three ways. First, CenterNet uses single-scale heatmaps, which are weak in terms of pose variation and occlusion. The *proposed multiscale*

*CenterNet*, as presented in Table 1, adopts multiscale heatmap generation. The predicted LR maps are used to learn the final HR maps more accurately. Table 1 confirms that this multiscale approach can overcome the drawbacks of the conventional CenterNet.

Second, the *proposed multiscale attention CenterNet* adds multiscale attention to CenterNet. In other words, internal LR features are jointly learned and fused with internal HR features based on scaled dot-product attention and between-hourglass skip connection. The internal LR features serve as a guide for learning the internal HR features. This enables internal LR and HR joint feature learning and increases the restoration accuracy of the HR heatmap. Owing to the LR and HR joint feature learning, the performance of the proposed multiscale attention CenterNet is improved, as presented in Table 1.

Third, the *proposed Mada-CenterNet* integrates deformable convolution into the between-hourglass skip connection to improve the proposed multiscale attention CenterNet. Unlike standard convolution, deformable convolution can enhance the CNN capability of modelling geometric deformations. The last row in Table 1 indicates that the proposed Mada-CenterNet can be more robust to occlusion and pose variation by determining more discriminative features, thereby improving the pest counting accuracy.

### 5.3. Pest counting results

Fig. 11 shows the detected bounding boxes with the conventional CenterNet[2], predicted density maps with KDMG [10], and detected bounding boxes with the proposed Mada-CenterNet. As presented in Table 1, KDMG and CenterNet were the best models among counting approaches based on density map estimation and detection, respectively. Therefore, only these two models were visually compared with the proposed Mada-CenterNet. In Fig. 11, the numbers above the images represent the measured and predicted pest count. As shown in the figures, the proposed model can provide a more accurate number of pests than the KDMG and CenterNet. Unlike the existing object detection datasets [5], the input pest images include severe pose and scale variations, occlusion, and color/shape similarities. Although the KDMG and CenterNet are SOTA models, they have limitations in overcoming challenging problems. However, the proposed Mada-CenterNet incorporates deformable and multiscale joint feature learning via the between-hourglass skip connection, thereby addressing occlusion and pose/scale variation problems, finally improving the pest counting accuracy.

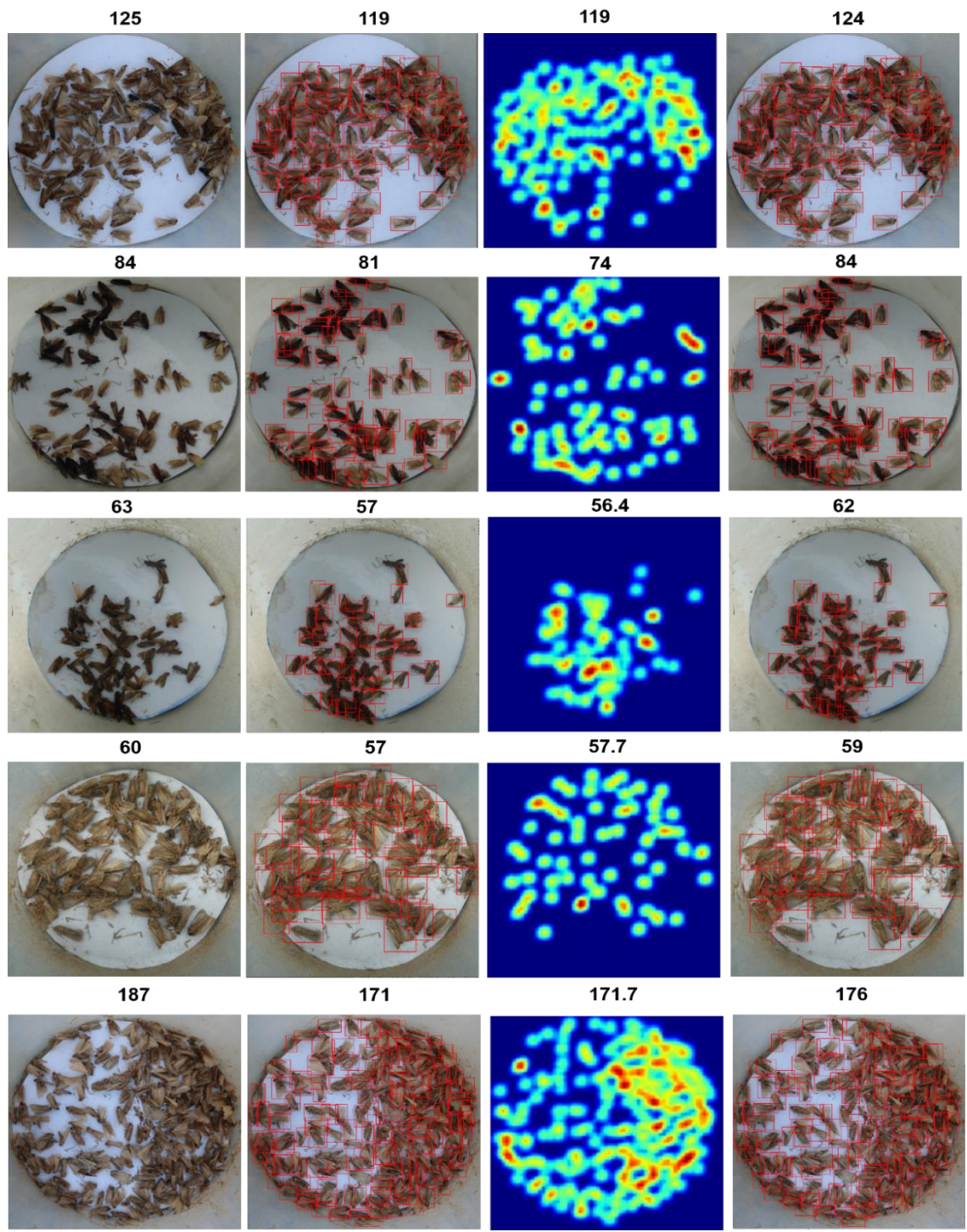

Fig. 11. Experimental results: input pest images, bounding boxes detected with CenterNet[2], density maps predicted with KDMG [10], and bounding boxes detected with the proposed Mada-CenterNet (left to right).

## 6. Conclusion

This study presents a new Mada-CenterNet model for trap-based pest counting. Unlike object detection image datasets, pest images captured by a lighting trap can have severe pose and scale variations and even occlusion. Moreover, they have a similar appearance in color and texture, making pest counting challenging. To solve these problems, three main aspects were advanced compared with CenterNet. First, a multiscale heatmap generation approach was adopted in a two-step fashion to adaptively learn the changes in the number of pests. Second, the internal LR and HR joint feature learning was modeled via the between-hourglass skip connection and scaled dot-product attention to boost the internal HR features. Third, geometric deformation was incorporated into the between-hourglass skip connection to be more agnostic to pose variations and occlusion problems. The experimental results confirmed that the proposed Mada-CenterNet was extremely effective in overcoming the scale/pose variations, texture similarity, and occlusion problems that appear in pest images. The proposed model upgraded the conventional CenterNet for pest counting. Moreover, the proposed Mada-CenterNet surpassed the SOTA models, including counting by detection and by density map estimation.

**Funding**

This work was carried out with the support of Cooperative Research Program for Agriculture Science & Technology Development (Grant no. PJ016303), National Institute of Crop Science (NICS), Rural Development Administration (RDA), Republic of Korea; National Research Foundation of Korea(NRF) grant funded by the Korea government(MSIT) (No. 2020R1A2C1010405).

**Acknowledgment**

We thank Dr. Hwijong Yi of the Rural Development Administration for his assistance in collecting pest images.

## References

[1] V. Lempitsky and A. Zisserman, "Learning to count objects in images," Advances in neural information

processing systems 23, pp. 1324-1332, Vancouver, Canada, 2010.

[2] X. Zhou, D. Wang and P. Krähenbühl, "Objects as points," arXiv:1904.07850v2 [cs.CV], Apr. 2019.

[3] Sh. Ren, K. He, R. Girshick and J. Sun, "Faster R-CNN: Towards real-time object detection with region proposal networks," IEEE Transactions on Pattern Analysis and Machine Intelligence, vol. 39, no. 6, pp. 1137-1149, June 2017.

[4] T. Lin, P. Goyal, R. Girshick, K. He and P. Dollár, "Focal loss for dense object detection," in Proc. IEEE International Conference on Computer Vision, pp. 2999-3007, Oct. 2017.

[5] T.-Y. Lin, M. Maire, S. Belongie, J. Hays, P. Perona, D. Ramanan, P. Dollar, and C. L. Zitnick, "Microsoft coco: Common objects in context," in Proc. European Conference on Computer Vision, Zurich, Switzerland, pp. 740-755, Sept. 2014.

[6] L. Boominathan, S. Kruthiventi, and R. V. Babu "CrowdNet: A deep convolution network for dense crowd counting," in Proc. ACM international conference on Multimedia, pp. 640-644, Aug. 2016.

[7] Y. Zhang, D. Zhou, S. Chen, S. Gao, and Y. Ma, "Single-image crowd counting via multi-column convolutional neural network," in Proc. IEEE Conference on Computer Vision and Pattern Recognition, Las Vegas, NV, USA, pp. 589-597, Jun. 2016.

[8] X. Cao, Z. Wang, Y. Zhao, and F. Su, "Scale aggregation network for accurate and efficient crowd counting," in Proc. European Conference on Computer Vision, Munich, Germany, pp. 757-773, Oct. 2018.

[9] V. Ranjan, H. Le, M. Hoai, "Iterative crowd counting," in Proc. European Conference on Computer Vision, Munich, Germany, pp. 278-293, Oct. 2018.

[10] J. Wan, Q. Wang, and A. B. Chan, "Kernel-based density map generation for dense object counting," IEEE Transactions on Pattern Analysis and Machine Intelligence, vol. 44, no. 3, March 2022.

[11] N. Dalal and B. Triggs, "Histograms of oriented gradients for human detection," in Proc. IEEE Conference on Computer Vision and Pattern Recognition, San Diego, CA, USA, June 2005, pp. 886-893.

[12] D. G. Lowe, "Distinct image features from scale-invariant key points," International Journal of Computer

Vision, vol. 60, pp. 91-110, 2004.

[13] C. Cortes and V. Vapnik, "Support-vector networks," Machine Learning, vol. 20, pp. 273-297, 1995.

[14] J. R. Uijlings, K. E. van de Sande, T. Gevers, A. W. Smeulders, "Selective search for object recognition," International Journal of Computer Vision, vol. 104, pp. 154–171, 2013.

[15] C. L. Zitnick and P. Dollár, "Edge Boxes: Locating object proposals from edges," in Proc. European Conference on Computer Vision, Zurich, Switzerland, Sept. 2014, pp. 391-405.

[16] J. Redmon, S. Divvala, R. Girshick, and A. Farhadi, "You only look once: Unified, realtime object detection," in Proc. IEEE Conf. Computer Vision and Pattern Recognition, Las Vegas, NV, USA, pp. 779-788, June 2016.

[17] W. Liu, D. Anguelov, D. Drhan, C. Szegedy, S. Reed, C.-Y. Fu, A. C. Berg, "SSD: single shot multibox detector," in Proc. European Conference on Computer Vision, Amsterdam, Netherlands, Oct. 2016, pp. 21-37.

[18] K. He, G. Gkioxari, P. Dollár, and R. Girshick, "Mask R-CNN," in Proc. International Conference on Computer Vision, Venice, Italy, Oct. 2017, pp. 2961-2969.

[19] Z. Cai and N. Vasconcelos, "Cascade R-CNN: Delving into high quality object detection," in Proc. IEEE Conference on Computer Vision and Pattern Recognition, Salt Lake City, USA, Jun. 2018, pp. 6154-6162.

[20] R. Kestur, A. Meduri, O. Narasipura, "MangoNet: A deep semantic segmentation architecture for a method to detect and count mangoes in an open orchard," Engineering Applications of Artificial Intelligence, vol. 77, pp. 59-69, Jan. 2019.

[21] D. Xia, P. Chen, B. Wang, J. Zhang and C. Xie, "Insect detection and classification based on an improved convolutional neural network," Sensors, vol. 18, no. 12, pp. 1-12, Nov. 2018.

[22] W. Ding and G. Taylor, "Automatic moth detection from trap images for pest management," Computers and Electronics in Agriculture, vol. 123, pp. 17-28, Apr. 2016.

[23] L. Liu, R. Wang, C. Xie, P. Yang, F. Wang, S. Sudirman, and W. Liu, "PestNet: An end-to-end deep learning approach for large-scale multi-class pest detection and classification," IEEE Access, vol. 7, pp. 45301-45312, Apr. 2019.

[24] D. P. Kingma and J. L. Ba, "Adam: A method for stochastic optimization," in Proc. International Conference on Learning Representations, San Diego, USA, May. 2015.

[25] A. Vaswani, N. Shazeer, N. Parmar, J. Uszkoreit, L. Jones, A. N. Gomez, Ł. Kaiser and I. Polosukhin, "Attention is All you Need," in Proc. Advances in Neural Information Processing Systems, 30, Long Beach, U.S.A., pp. 5998-6008, Dec. 2017.

[26] A. Dosovitskiy, L. Beyer, A. Kolesnikov, D. Weissenborn, X. Zhai, T. Unterthiner, M. Dehghani, M. Minderer, G. Heigold, S. Gelly, J. Uszkoreit and N. Houlsby, "An image is worth 16x16 words: Transformers for image recognition at Scale," arXiv:2010.11929[cs.CV], Oct. 2020.

[27] W. Wang, E. Xie, X. Li, D. P. Fan, K. Song, D. Liand, T. Lu, P. Luo and L. Shao, "Pyramid vision transformer: A versatile backbone for dense prediction without convolutions," in Proc. IEEE/CVF International Conference on Computer Vision, 2021, pp. 548-558.

[28] Z. Yang, S. Liu, H. Hu, L. Wang and S. Lin, "RepPoints: Point set representation for object detection," in Proc. IEEE International Conference on Computer Vision, pp. 9657-9664, Aug. 2019.